\documentclass[pdflatex,sn-mathphys-num]{sn-jnl}% Math and Physical Sciences 

\usepackage{graphicx}%
\usepackage{multirow}%
\usepackage{amsmath,amssymb,amsfonts}%
\usepackage{amsthm}%
\usepackage{mathrsfs}%
\usepackage[title]{appendix}%
\usepackage{xcolor}%
\usepackage{textcomp}%
\usepackage{manyfoot}%
\usepackage{booktabs}%
\usepackage{float}
\usepackage{float}
\usepackage[ruled,vlined]{algorithm2e}
\usepackage{adjustbox}

\theoremstyle{thmstyleone}%
\newtheorem{theorem}{Theorem}%  meant for continuous numbers.
\theoremstyle{thmstyleone}

\newtheorem{lemma}[theorem]{Lemma}

\theoremstyle{thmstyletwo}%

\theoremstyle{thmstylethree}%

\begin{document}

\title[Article Title]{Balancing Safety and Optimality in Robot Path Planning: Algorithm and Metric}

\author*[1,]{\fnm{Jatin Kumar} \sur{Arora}}\email{arorajatinkumar2000@gmail.com}

\author[1]{\fnm{Soutrik} \sur{Bandyopadhyay}}\email{	Soutrik.Bandyopadhyay@ee.itd.ac.in}

\author[2]{\fnm{Sunil} \sur{Sulania}}\email{sulaniasunil@gmail.com}

\author[1]{\fnm{Shubhendu} \sur{Bhasin}}\email{Shubhendu.Bhasin@ee.iitd.ac.in}

\affil*[1]{\orgdiv{Department of Electrical Engineering}, \orgname{Indian Institute of Technology Delhi}, \orgaddress{\city{Delhi}, \country{India}}}

\affil[2]{\orgname{Samsung R\&D Institute}, \orgaddress{\city{Delhi}, \country{India}}}

\abstract{Path planning for autonomous robots faces a fundamental trade-off between path length and obstacle clearance. While existing algorithms typically prioritize a single objective, we introduce the Unified Path Planner (UPP), a graph-search algorithm that dynamically balances safety and optimality via adaptive heuristic weighting. UPP employs a local inverse-distance safety field and auto-tunes its parameters based on real-time search progress, maintaining bounded heuristic behavior while maintaining superior clearance. To enable rigorous evaluation, we introduce the OptiSafe index, a normalized metric that quantifies the trade-off between safety and optimality. Extensive evaluation across 10 environments shows that UPP achieves a 0.94 OptiSafe score in cluttered environments, compared with 0.22-0.85 for existing methods, with only 0.5-1\% path-length overhead in simulation and a 100\% success rate. Hardware validation on TurtleBot confirms practical advantages despite sim-to-real gaps.}


\maketitle

\section{Introduction}
\label{sec:intro}
The increasing integration of robots into complex human environments has made
robust and efficient path planning a cornerstone of modern robotics. To ensure
smooth, safe, and productive operation from autonomous vehicles \cite{Paden2016,
  ChaochengLi2015, Lee2014} and UAVs \cite{Bortoff2000} to robotic arms
\cite{Dai2022, Korayem2011, Kim1985, Klanke2006} and warehouse automation
systems \cite{Hvezda2018, Vivaldini2010}, robots must navigate dynamic and
cluttered spaces effectively. To navigate effectively in the real world,
planners should find safe paths without compromising optimality. However,
current planners focus either on optimality or safety, prioritizing one at the
expense of the other. Finding a path that is both safe and near optimal remains
an unresolved challenge in cluttered environments. 

Path planning algorithms span multiple families, each with distinct trade-offs.
Grid-based search algorithms, such as A* \cite{Hart1968}, form a foundational
category. Innovations in this area include modifications to accelerate search
time at the cost of optimality \cite{Warren1993} and techniques for planning
over continuous grid edges to produce shorter paths \cite{Nash2013}. For dynamic
environments, replanning algorithms like D* \cite{Stentz1994} have been
developed to efficiently update paths without recalculating from scratch.
Sampling-based algorithms like the Rapidly Exploring Random Tree (RRT)
\cite{Noreen2016} have revolutionized planning in high-dimensional configuration
spaces. Subsequent variants, such as RRTSmart \cite{Nasir2013}, introduced
intelligent sampling and path optimization to improve convergence and cost,
while FASTRRT \cite{Wu2021} significantly reduced execution time through
advanced sampling and steering strategies. Combinatorial methods \cite{Lav06,
  Jafarzadeh2018, Ayawli2019}, such as those based on Voronoi diagrams
\cite{Ayawli2019}, focus on maximizing path safety by maintaining the largest
possible distance from obstacles. Furthermore, potential field-based methods
\cite{Hwang1992, Rostami2019, Azzabi2019} draw inspiration from physics, using
attractive and repulsive forces to guide the robot. Modified versions of these
algorithms have successfully addressed classic problems, such as avoiding local
minima. 

These algorithms excel at single objectives,  A* variants optimize path length,
while Voronoi-based methods maximize clearance; but lack adaptive mechanisms to
jointly balance them. Existing hybrid approaches suffer from (1) hand-tuned
parameters requiring manual adjustment per environment, (2) excessive
conservatism in cluttered spaces, or (3) no theoretical guarantees on path cost.
Recent research has made significant strides toward bridging this gap.
\cite{Chu2024} proposed Optimized A*, an improved A* with turning and obstacle
penalty terms, but its handcrafted cost may sacrifice path optimality for small
safety gains or still cut too close to obstacles in cluttered, narrow regions.
Control Barrier Functions (CBFs) have been integrated into planners like CBF-RRT
\cite{Manjunath2021} to enforce safety-critical constraints, demonstrating
superior performance over other RRT-based methods in generating safe paths. It
ensures safety but struggles with exploration under CBF constraints. SDF A*
\cite{Wang2023} leverages a Signed Distance Function (SDF) to embed safety
information directly into the planning process, showing significant improvements
in both safety and exploration compared to traditional algorithms, but it
becomes overly conservative. FS Planner \cite{Cobano2025} uses an integrated
inverse distance between nodes on lazy theta A* and also gives a limit on the
maximum neighbors explored in the goal and safe direction. However, this bias
can miss good paths in tight passages and offers little guarantee of
near-optimal path cost. Despite these advancements, existing solutions that
address both objectives often come with significant drawbacks. They can incur a
large trade-off that excessively penalizes path cost for minor gains in safety,
or rely on fixed, non-tunable safety margins that cannot adapt to changing
operational needs.

Another limitation in existing research is the evaluation methodology used by
path planners. Most studies report performance using isolated metrics, such as
path length or minimum clearance. While individually informative, these metrics
do not capture the underlying trade-off between safety and optimality, allowing
planners to appear effective by excelling in one dimension while performing
poorly in the other. This fragmentation makes it challenging to compare
safety-aware planners fairly or understand how algorithmic choices influence
real-world safety.  

To address the fundamental challenge of balancing safety and optimality in
robotic path planning, we present an integrated framework that spans planning
and evaluation. At its core, we introduce the Unified Path Planner (UPP), a
safety-aware graph search algorithm that embeds obstacle proximity directly into
the heuristic via a locally computed inverse-distance safety field, enabling
simultaneous optimization of safety and path length. Crucially, unlike existing
safety-aware planners that rely on fixed, hand-tuned parameters or conservative
inflation schemes, to the best of our knowledge, UPP is the first to combine
online auto-tuning of the safety–optimality trade-off while  maintaining bounded heuristic behavior, ensuring theoretical bounds with adaptivity. As balancing
optimality and safety is essential in robotics, we propose the OptiSafe Index.
This unified and normalized metric jointly measures the balance and strength of
safety and optimality in a single score. Extensive simulation studies and
real-world TurtleBot experiments demonstrate that this combined approach reveals
safety-optimality trade-offs that conventional planners typically do not
consider.  

It is important to clarify that the objective of UPP is not to maximize
obstacle clearance unconditionally. In a deterministic environment, once
an application-dependent safety margin is satisfied, further increases in
clearance may provide limited benefit while unnecessarily increasing path
length. Instead, UPP treats obstacle clearance as a soft safety preference
and seeks increased separation from obstacles while maintaining a
near-optimal path. This distinction is reflected in the adaptive safety
weight, which reduces the influence of the safety term when it impedes
progress toward the goal. Thus, UPP is intended to avoid both
obstacle-proximal shortest paths and excessively conservative high-clearance
paths, providing a compromise between the two objectives.

\section{Preliminaries}

For the planning problem, let us consider \( G \) to be an undirected graph with
the set of nodes
\[
N \triangleq \{ n_i \mid i = 1,2,\dots,K \}, \qquad K \in \mathbb{N},
\]
where each node state $n_i$ is  
\[
n_i =
\begin{cases}
0, & \text{free}, \\
\neq 0 & \text{occupied or undetermined}.
\end{cases}
\]

\noindent We denote the set of edges for the undirected graph $G$ as
\[
E \triangleq \{ c_{ij} \mid i,j \in N \},
\]
where \( c_{ij} \) denotes the cost of the directed edge from \( i \) to \( j \) and, since the graph is undirected,
\[
c_{ij} = c_{ji}, \qquad c_{ij} > 0.
\]
We assume there exists a constant \( \delta > 0 \) such that
\[
c_{ij} \ge \delta \quad \forall (i,j) \in E.
\]

Let \( \Gamma : N \to \Gamma(N) \) be the successor  mapping, where  
\( \Gamma(N) \) denotes the set of neighbors of a node.
We denote the starting node and the preferred goal node by $s \in N$ and $t \in N$
respectively. Let the cost functions be
\[
g(n):  \mathbb{R}^n \to \mathbb{R}_+  \text{: Cost from node } s \text{ to node } n, \qquad
h(n):  \mathbb{R}^n \to \mathbb{R}_+  \text{: Cost from node } n \text{ to node } t, 
\]
and the evaluation function 
\[
f(n) \triangleq g(n) + h(n),
\]
which is to be minimized.

The objective of the planner is to find a sequence of nodes.
\[
\pi = (n_0, n_1, \dots, n_L), \qquad n_0 = s,\; n_L = t,
\]
such that each consecutive pair satisfies \( (n_k, n_{k+1}) \in E \), and the total accumulated path cost
\[
J(\pi) = \sum_{k=0}^{L-1} c_{n_k n_{k+1}}, 
\]
is minimized. The evaluation function provides an estimate of the total path cost through \( n \) and determines the node-expansion order during the search.

For any node \( n \), the successor set \( \Gamma(n) \subseteq N \) denotes the set
of nodes reachable from \( n \) via feasible edges. In grid-based planning, this
typically corresponds to the set of adjacent lattice cells (e.g., 4, 8, or
26-connected neighborhoods), whereas in arbitrary graphs it denotes the
immediate neighbors with finite edge costs. 

From \cite{Hart1968}, we additionally define the optimal cost-to-go function,
\[
h^\ast(n) = \min_{\pi : n \to t} J(\pi),
\]
which represents the true minimal cost from \( n \) to the goal node t. A heuristic \( h(n) \) is called \emph{admissible}  if
\[
0 \le h(n) \le h^\ast(n), \quad \forall n \in N,
\]
and \emph{consistent} if it satisfies
\[
h(n) \le c_{n n'} + h(n'), \qquad \forall n' \in \Gamma(n).
\]

Finally, the search process maintains an OPEN set of frontier nodes ordered by
increasing \( f \)-value and a CLOSED set of expanded nodes. At each iteration,
the node with minimal \( f(n) \) is selected for expansion. The process
terminates either when the goal node \( t \) is extracted from OPEN or when OPEN
becomes empty, in which case no feasible path exists. In the rest of the paper,
optimality is measured by path length, while safety is measured by minimum
clearance. 

\section{Unified Path Planner (UPP)}
\label{sec:upp}
The Unified Path Planner (UPP) is a graph search algorithm that generates safe,
efficient, and smooth paths by combining geometric distance cues with a locally
computed obstacle-proximity field. Unlike planners that rely on fixed, manually
tuned heuristic weights, UPP incorporates an adaptive mechanism that
continuously adjusts its heuristic parameters based on real-time feedback during
the search.

The UPP employs a heuristic that combines geometric
distance-to-go with a locally computed safety potential, enabling the
planner to simultaneously account for path efficiency and obstacle
avoidance. For any node \( n \in N \), the heuristic is defined as
\begin{equation}
    h(n) \;=\;
    \alpha\, l_{1}(n,t)
    + (1-\alpha)\, l_{\infty}(n,t)
    + \beta\, S(n),
    \label{eq:upp_heuristic}
\end{equation}
where \( \alpha \in [0,1] \) and \( \beta > 0 \) are adaptive weighting
parameters. The terms \( l_1 \) and \( l_{\infty} \) correspond to the
Manhattan and Chebyshev distances.
$l_{1}(n,t) = \lVert n - t \rVert_{1}$ and $l_{\infty}(n,t) = \lVert n - t \rVert_{\infty}$.

\begin{figure}[t]
    \centering
    \includegraphics[width=0.55\linewidth]{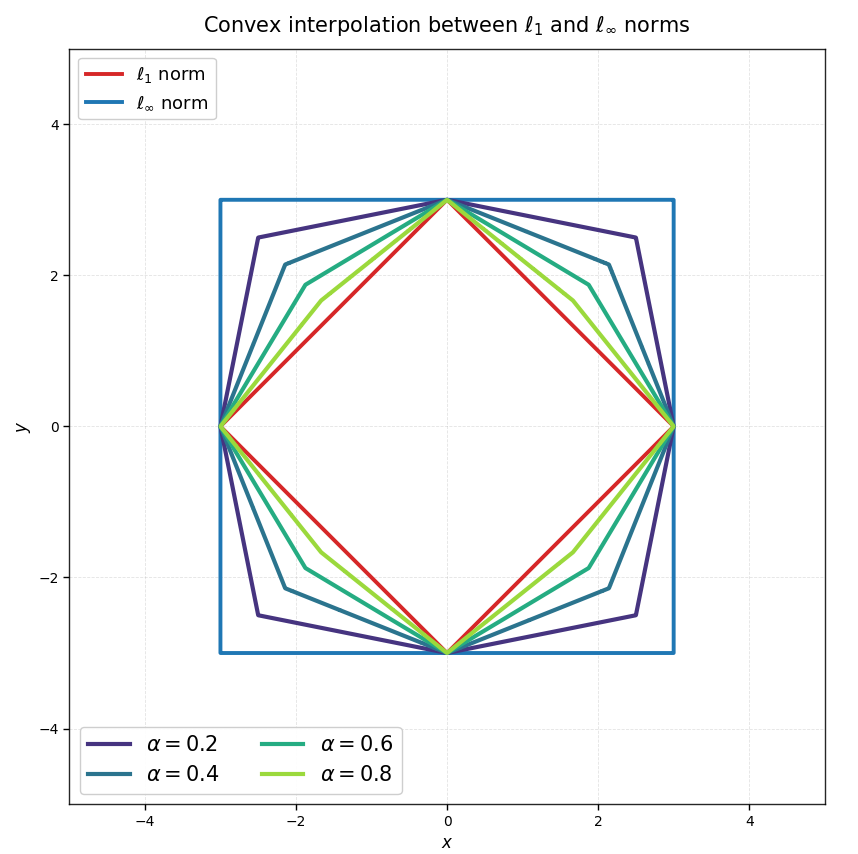}
    \caption{
    Level set visualization of the $\ell_1$ norm , the $\ell_\infty$ norm ,
    and their convex combinations
    for multiple values of $\alpha$.
}
    \label{fig:norm_combined}
\end{figure}

The choice of combining the Manhattan and Chebyshev distances is motivated by their distinct geometric structures. The level sets of the Manhattan distance form diamond-shaped contours aligned with the grid axes, encouraging diagonal progression when minimizing cost. In contrast, the level sets of the Chebyshev distance form axis-aligned squares, which promote straight-line motion along the horizontal and vertical directions. As illustrated in Fig.~\ref{fig:norm_combined}, the color-coded level sets reveal that the combined heuristic inherits desirable properties from both norms, enabling smoother, more natural directional transitions during planning.

\begin{figure}[t]
    \centering
    \includegraphics[width=0.75\linewidth]{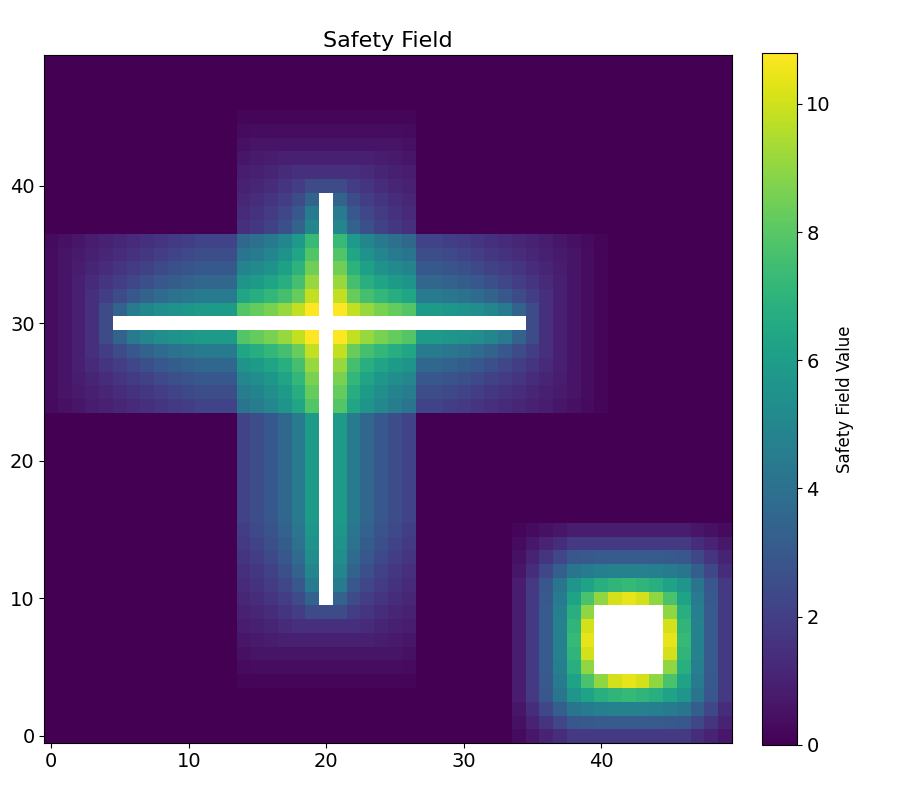}
    \caption{
        Safety Field on a grid, white area represents obstacles
    }
    \label{fig:safety_field}
\end{figure}

The term \( S(n) \) represents a safety potential encoding the proximity
of node \( n \) to surrounding obstacles. It is defined over a Chebyshev-radius neighborhood of size \( R \) as
\begin{equation}
    S(n)
    = \!\!\!\sum_{\Delta \in [-R,R]}
    \frac{\mathbf{1}_\mathcal{O}(n + \Delta)}
         {\|\Delta\|_{\infty} + \varepsilon},
    \label{eq:safe}
\end{equation}
where  $\mathbf{1}_\mathcal{O}(n + \Delta)$ is the indicator
function, equal to $\mathrm{1}$ if the grid cell at $n+\Delta$ is occupied by an
obstacle and $0$ otherwise,
\( \mathcal{O} \subset N \) denotes the set of obstacle nodes,
and \( \varepsilon > 0 \) prevents division singularity. This formulation
implements an inverse-distance accumulation as nearby obstacles contribute
larger penalties, while distant obstacles within the safety horizon make
smaller contributions. The safety field is efficiently computed via
convolution using a preconstructed kernel
\[
    K(\Delta) = 
    \begin{cases}
        \dfrac{1}{\|\Delta\|_{\infty} + \varepsilon},
        & 1 \le \|\Delta\|_{\infty} \le R, \\[6pt]
        0, & \text{otherwise},
    \end{cases}
\]
ensuring that \( S(n) \) is available in constant time during the search. The safety heuristic for a small grid for $R=6$  is illustrated in Figure ~\ref{fig:safety_field}

Unlike classical inflation-based heuristics that uniformly pad obstacle
regions, the inverse distance safety potential offers a continuous and
geometry-aware measure of local risk. This enables the planner to adapt
its behavior seamlessly between open, low-risk areas and narrow, obstacle-dense corridors.

\subsection{Grid Connectivity}

The planner operates on an $8$-connected occupancy grid. Each node is connected
to its horizontal, vertical, and diagonal neighbors.

The edge cost between neighboring nodes is defined as the Euclidean distance

\[
c(n,m)=\|m-n\|_2.
\]

Consequently,

\[
c(n,m)=
\begin{cases}
1,&\text{horizontal or vertical move},\\
\sqrt{2},&\text{diagonal move}.
\end{cases}
\]

All heuristic formulations and theoretical analysis in this chapter are based on this connectivity and edge-cost model.

\subsection{Initial Parameter Selection}
\label{subsec:upp_init_params}

The Unified Path Planner uses four main heuristic parameters:
the distance-mixing weight \(\alpha\), the safety weight \(\beta\),
the safety radius \(R\), and a small constant \(\varepsilon\) used for
numerical stability.

Initially, the distance-mixing weight \(\alpha\) is set to
\(\alpha = 0.5\), which gives an equal balance between the Manhattan
and Chebyshev distance terms in the heuristic. This neutral choice
avoids biasing the planner toward either purely axis-aligned or purely
diagonal motion at the outset. During the search, \(\alpha\) is adapted
based on the accumulated turning behavior of the path. 

The constant \(\varepsilon > 0\) appears in the inverse distance safety
field to prevent division by zero when an obstacle lies directly
adjacent to a node. In practice, a small value such as
\(\varepsilon = 0.01\) is sufficient to regularize the safety term
without noticeably affecting its magnitude.

The initial safety weight \(\beta\) and safety radius \(R\) are computed
from the geometry of the occupancy grid. Let
\(D = \operatorname{DT}(G)\) denote the Euclidean distance transform of
the free space, computed as
\[
    D(x) = \text{dist}(x,\mathcal{O}), \quad
    \mathcal{O} = \{ x \mid G(x) = 1 \},
\]
and let \(\mathcal{F} = \{ x \mid G(x) = 0 \}\) be the set of free cells.
We define
\[
    \mu = \mathbb{E}[D(x) \mid x \in \mathcal{F}], \qquad
    \sigma = \operatorname{Std}[D(x) \mid x \in \mathcal{F}],
\]
as the mean and standard deviation of the free space distance field, and
\[
    \rho = \frac{|\mathcal{O}|}{|G|}
\]
as the obstacle density, i.e., the fraction of occupied cells in the
grid. Intuitively, larger \(\rho\) and larger relative variability
\(\sigma / (\mu + \varepsilon)\) indicates more cluttered or irregular
environments, in which safety should be weighted more strongly.

Given a user-specified base value \(\beta\), the initial safety weight is
scaled according to
\begin{equation*}
    \beta_{\text{init}}
    \;=\;
    \operatorname{clip}
    \!\left(
        \frac{\beta \rho \sigma}{\mu + \varepsilon},
        \,\beta_{\min},\,\beta_{\max}
    \right),
    \label{eq:beta_init}
\end{equation*}
where \(\operatorname{clip}(\cdot,\beta_{\min},\beta_{\max})\) saturates
the value to lie within predefined bounds
\([\beta_{\min},\beta_{\max}]\). This mapping increases \(\beta\) in
dense, highly variable environments and attenuates it in sparse, more
uniform maps, thereby adapting the safety influence to the overall
structure of the workspace.

Similarly, the initial safety radius \(R\) is rescaled using the same
statistics:
\begin{equation*}
    R_{\text{init}}
    \;=\;
    \operatorname{clip}
    \!\left(
        \operatorname{round}\big(R\,(\mu + \sigma)\big),
        \,R_{\min},\,R_{\max}
    \right),
    \label{eq:R_init}
\end{equation*}

Larger values of \(\mu + \sigma\) typically correspond to more spacious
environments, for which a larger lookahead radius is beneficial, whereas
tighter or more cluttered maps keep \(r\) small to control computational
cost. 

\subsection{Adaptation of Parameters}
\label{subsec:upp_adaptivity}

UPP incorporates online parameter adaptation for both the safety weight
\(\beta\) and the distance mixing parameter \(\alpha\), enabling the
heuristic to respond dynamically to local geometry and search progress.
These updates operate after the geometry-aware initialization described
in Algorithm~\ref{alg:upp_init} and are applied during node expansion.

Let \(n_i\) denote the node expanded at iteration \(i\), and let
\[
d_i \triangleq \|n_i - t\|_2
\]
be its Euclidean distance to the goal. We define the signed progress.
\[
\Delta d_i \triangleq d_i - d_{i-1}.
\]
Given a tolerance \(\tau_{\text{goal}}>0\), we define a progress indicator \(p_i\) as:
\[
p_i =
\begin{cases}
+1, & \Delta d_i < -\tau_{\text{goal}} \quad\text{, progress}\\
-1, & \Delta d_i > \tau_{\text{goal}} \quad\text{, regression}\\
0,  & |\Delta d_i|\le \tau_{\text{goal}} \quad\text{, stall}.
\end{cases}
\]

A stall counter \(s_i\) is maintained as
\[
s_i =
\begin{cases}
0, & p_i \neq 0,\\
s_{i-1}+1, & p_i = 0,
\end{cases}
\qquad s_0 = 0.
\]
Let \(\gamma_{\text{rec}}>1\) and \(\gamma_{\text{dec}}\in(0,1)\) denote
multiplicative recovery and decay factors, and let
\([\beta_{\min},\beta_{\max}]\) be fixed bounds. The safety weight is
updated according to
\begin{equation}
\beta_{i+1}
=
\Pi_{[\beta_{\min},\,\beta_{\max}]}
\left(
\beta_i\,
\gamma_{\mathrm{rec}}^{\mathbf{1}_{\{p_i=+1\}}}
\gamma_{\mathrm{dec}}^{\mathbf{1}_{\{p_i=-1\}} + \mathbf{1}_{\{s_i \ge K_\beta\}}}
\right).
\label{eq:beta_update}
\end{equation}

where \(\Pi_{[a,b]}(\cdot)\) denotes projection onto \([a,b]\),
\(\mathbf{1}_{\mathcal{A}}\) is the indicator function in set A, and \(K_\beta\) is a
patience threshold. This rule reduces \(\beta\) when the search stalls or
moves away from the goal, allowing escape from overly conservative
regions, while increasing \(\beta\) during sustained progress to
encourage safer exploration.

To regulate path smoothness, UPP adapts the distance mixing parameter
\(\alpha\) based on turning behavior. Let \(p(n_i)\) denote the parent of
\(n_i\), and define the motion and goal direction vectors
\[
v_i = n_i - p(n_i),
\qquad
g_i = t - n_i.
\]
Let \(\theta_i^{\text{move}}\) and \(\theta_i^{\text{goal}}\) be their
headings, and define the wrapped angular deviation
\[
\theta_i \triangleq
\left|
\mathrm{wrap}_{[-\pi,\pi]}
\!\left(
\theta_i^{\text{goal}} - \theta_i^{\text{move}}
\right)
\right|.
\]
where $\mathrm{wrap}_{[-\pi,\pi]}(\alpha) \triangleq ((\alpha+\pi)\bmod 2\pi)-\pi$.

Over a window of length \(K_\alpha\), UPP accumulates deviation relative
to a target turning angle \(\theta_{\text{tar}}\):
\[
u_i \triangleq
\sum_{j=i-K_\alpha+1}^{i}
(\theta_j - \theta_{\text{tar}}).
\]
With hysteresis threshold \(\tau_{\text{ang}}>0\), recovery factor
\(\eta_{\text{rec}}>1\), decay factor \(\eta_{\text{dec}}\in(0,1)\), and
bounds \([\alpha_{\min},\alpha_{\max}]\), the update rule is
\begin{equation}
\alpha_{i+1} =
\begin{cases}
\Pi_{[\alpha_{\min},\alpha_{\max}]}
(\alpha_i\,\eta_{\text{rec}}), & u_i > \tau_{\text{ang}},\\[4pt]
\Pi_{[\alpha_{\min},\alpha_{\max}]}
(\alpha_i\,\eta_{\text{dec}}), & u_i < -\tau_{\text{ang}},\\[4pt]
\alpha_i, & \text{otherwise}.
\end{cases}
\label{eq:alpha_update}
\end{equation}

Increasing \(\alpha\) strengthens the Manhattan component of the hybrid
distance, discouraging diagonal motion, while decreasing \(\alpha\)
permits greater diagonal freedom via the Chebyshev component.

Global map statistics provide a
well-scaled prior, while online adaptivity balances safety, optimality,
and smoothness as the search evolves.
\subsection{Handling Heuristic Changes During Search}
\label{subsec:priority_handling}

Since the parameters $\alpha$ and $\beta$ are adapted during the search, the
heuristic function is time-varying. Let $\alpha_k$ and $\beta_k$ denote the
parameter values at search iteration $k$. The heuristic used at that iteration is
\[
h_k(n)
=
\alpha_k \ell_1(n,t)
+
(1-\alpha_k)\ell_\infty(n,t)
+
\beta_k S(n).
\]

In the present implementation, the priority of a node is computed when the node
is inserted into the OPEN list. If $\alpha$ or $\beta$ is subsequently updated,
nodes already contained in OPEN retain their previously computed priority values.
The updated parameter values are used only for heuristic evaluations performed
after the update, particularly for newly generated neighboring nodes or nodes
inserted again following an improvement in their cost-to-come.

Therefore, the OPEN list may temporarily contain nodes whose priority values were
computed using different historical values of $\alpha$ and $\beta$. The entire
priority queue is not rebuilt after every parameter update, since doing so would
introduce additional computational overhead. Similarly, nodes already placed in
the CLOSED set are not reopened solely because of a change in the heuristic
parameters.

\subsection{Theoretical Properties}

\subsubsection{Theoretical Assumptions}

The theoretical results presented in this section are established under the following assumptions.

\begin{enumerate}
    \item \textbf{Static Environment:}
    The planning environment is represented by a static occupancy grid throughout the search. Obstacle locations remain fixed during the execution of the planner.

    \item \textbf{Finite Graph:}
    The occupancy grid induces a finite graph $G=(N,E)$ consisting of a finite number of nodes and edges.

    \item \textbf{Positive Edge Costs:}
    Every traversable edge satisfies
    \[
        c_{ij} \ge \delta > 0,
    \]
    where $\delta$ is a positive constant. Consequently, every feasible path has finite positive cost.

    \item \textbf{Bounded Adaptive Parameters:}
    The adaptive heuristic parameters are projected onto compact intervals according to the update laws described in Section~\ref{subsec:upp_adaptivity}. Specifically,
    \[
        \alpha \in [\alpha_{\min},\alpha_{\max}] \subset (0,1),
    \]
    \[
        \beta \in [\beta_{\min},\beta_{\max}],
    \]
    and the safety radius satisfies
    \[
        R \in [R_{\min},R_{\max}].
    \]
    Therefore, all adaptive parameters remain uniformly bounded throughout the search.

    \item \textbf{Deterministic Search:}
    Node expansion follows the evaluation function
    \[
        f(n)=g(n)+h(n),
    \]
\end{enumerate}
\subsubsection{Completeness of UPP}
\begin{lemma}
\label{lemma:bound}
Consider a finite $q$-dimensional occupancy grid using the full local
neighborhood, with axial edge cost $1$ and Euclidean costs for diagonal
movements. Suppose that the adaptive parameters satisfy
\[
\alpha \in [\alpha_{\min},\alpha_{\max}] \subset (0,1),
\qquad
\beta \in [\beta_{\min},\beta_{\max}],
\]
and that the safety radius satisfies
\[
R \in [R_{\min},R_{\max}].
\]

Then, for the two-dimensional $8$-connected motion model, the UPP heuristic
\[
h(n)
=
\alpha \ell_1(n,t)
+
(1-\alpha)\ell_\infty(n,t)
+
\beta S(n)
\]
is bounded by
\begin{equation}
\label{eq:heuristic_bound}
h(n)
\le
\sqrt{2}\,h^*(n)
+
\beta_{\max}
\sum_{d=1}^{R}
\frac{(2d+1)^q-(2d-1)^q}{d+\varepsilon},
\end{equation}
where $q$ denotes the dimension of the grid, $R$ denotes the finite
support radius of the obstacle-risk kernel, and $\varepsilon>0$ is the
regularization parameter defined in~\eqref{eq:safe}.
\end{lemma}

\begin{proof}
The planner is initialized with user-specified base parameters
$(\alpha_0,\beta_0,R)$. Before planning, $\beta$ and $R$ are rescaled
using statistics of the free-space distance transform and projected onto the
compact intervals
\[
[\beta_{\min},\beta_{\max}]
\qquad\text{and}\qquad
[R_{\min},R_{\max}],
\]
respectively. Similarly, the distance-mixing parameter $\alpha$ is restricted
to the compact interval
\[
[\alpha_{\min},\alpha_{\max}]\subset(0,1).
\]

During the search, $\alpha$ and $\beta$ are updated according to
\eqref{eq:alpha_update} and \eqref{eq:beta_update}. Since both update laws
include projection onto their respective admissible intervals,
\[
\alpha_{\min}\le\alpha\le\alpha_{\max},
\qquad
\beta_{\min}\le\beta\le\beta_{\max}.
\]

The obstacle-risk potential $S(n)$ is computed by convolving the binary
obstacle indicator with a finite-support inverse-distance kernel. For a kernel
radius $r$, obstacle cells contribute only when their Chebyshev distance from
$n$ satisfies
\[
1\le \|m-n\|_\infty\le R.
\]

In a $q$-dimensional grid, the number of cells at exact Chebyshev distance
$d$ from a node is
\[
k_d=(2d+1)^q-(2d-1)^q.
\]

Therefore, the obstacle-risk potential satisfies
\[
0\le S(n)\le S_{\max},
\]
where
\[
S_{\max}
=
\sum_{d=1}^{R}
\frac{k_d}{d+\varepsilon}
=
\sum_{d=1}^{R}
\frac{(2d+1)^q-(2d-1)^q}{d+\varepsilon}.
\]

It remains to bound the geometric component of the heuristic. For the
two-dimensional $8$-connected grid with axial edge cost $1$ and diagonal edge
cost $\sqrt{2}$, let
\[
\Delta x = |x_n-x_t|,
\qquad
\Delta y = |y_n-y_t|.
\]
Without loss of generality, assume that $\Delta x\ge\Delta y$. The Manhattan
and Chebyshev distances are then
\[
\ell_1(n,t)=\Delta x+\Delta y,
\qquad
\ell_\infty(n,t)=\Delta x.
\]

The obstacle-free shortest-path cost under this motion model is the octile
distance
\[
d_{\mathrm{oct}}(n,t)
=
(\Delta x-\Delta y)+\sqrt{2}\,\Delta y.
\]
Since obstacles cannot reduce the true optimal cost-to-go,
\[
d_{\mathrm{oct}}(n,t)\le h^*(n).
\]

Furthermore,
\[
\ell_1(n,t)
=
\Delta x+\Delta y
\le
\sqrt{2}
\left[
(\Delta x-\Delta y)+\sqrt{2}\,\Delta y
\right].
\]
Hence,
\[
\ell_1(n,t)
\le
\sqrt{2}\,d_{\mathrm{oct}}(n,t)
\le
\sqrt{2}\,h^*(n).
\]

Since
\[
\ell_\infty(n,t)\le \ell_1(n,t)
\]
and $0\le\alpha\le1$, the mixed geometric term satisfies
\[
\begin{aligned}
\alpha\ell_1(n,t)
+
(1-\alpha)\ell_\infty(n,t)
&\le
\alpha\ell_1(n,t)
+
(1-\alpha)\ell_1(n,t)\\
&=
\ell_1(n,t)\\
&\le
\sqrt{2}\,h^*(n).
\end{aligned}
\]

Using $\beta\le\beta_{\max}$ and $S(n)\le S_{\max}$, we obtain
\[
\begin{aligned}
h(n)
&=
\alpha\ell_1(n,t)
+
(1-\alpha)\ell_\infty(n,t)
+
\beta S(n)\\
&\le
\sqrt{2}\,h^*(n)
+
\beta_{\max}S_{\max}\\
&\le
\sqrt{2}\,h^*(n)
+
\beta_{\max}
\sum_{d=1}^{R}
\frac{(2d+1)^q-(2d-1)^q}{d+\varepsilon}.
\end{aligned}
\]

Thus, the UPP heuristic remains bounded throughout the search under the stated
parameter and motion-model assumptions.
\end{proof}

\begin{lemma}
Let $n$ be the currently expanded node and let $m_i, m_j \in \mathcal{N}(n)$ be two
neighbors. Suppose the UPP heuristic is
\[
h(m) = d(m,t) + \beta S(m),
\]
where $d(\cdot,t)$ is the distance heuristic and $S(\cdot)$ is the safety score,
bounded as $0 \le S(m) \le S_{\max}$.  

If
\[
|d(m_i,t) - d(m_j,t)| \le c
\quad \text{and} \quad
\beta |S(m_i) - S(m_j)| > c,
\]
then
\[
h(m_i) < h(m_j)
\iff
S(m_i) > S(m_j).
\]
In particular, when the current node is sufficiently far from the goal so that
distance variations among neighbors are bounded by $c$, the heuristic ordering is
dominated by safety differences.
\end{lemma}

\begin{proof}
For any two neighbors $m_i, m_j$ of $n$, the heuristic difference satisfies
\[
h(m_i) - h(m_j)
= \bigl[d(m_i,t) - d(m_j,t)\bigr]
+ \beta \bigl[S(m_i) - S(m_j)\bigr].
\]

When $n$ is far from the goal, the grid structure implies that the one-step distance
variations are bounded, i.e.,
\[
|d(m_i,t) - d(m_j,t)| \le c,
\]
for some constant $c > 0$.  

If $\beta |S(m_i) - S(m_j)| > c$, the safety term strictly dominates the distance
term, and therefore the sign of $h(m_i) - h(m_j)$ is determined by
$S(m_i) - S(m_j)$. Hence,
\[
h(m_i) < h(m_j) \iff S(m_i) > S(m_j),
\]
so the neighbor with higher safety is preferred.

As $n$ approaches the goal, the variation in $d(\cdot,t)$ among neighbors becomes
comparable to or larger than the safety variation, and the distance term governs
the ordering. Thus, UPP prioritizes safety when safety differences are meaningful,
and reverts to distance-based ordering near the goal.
\label{lemma:safety}
\end{proof}
\begin{theorem}
If a collision-free path from $s$ to $t$ exists in $G$, then UPP returns a feasible collision-free path whenever one exists, while favoring safety. Otherwise, it correctly reports failure. 
\end{theorem}

\begin{proof}
As the heuristics cost is bounded according to Lemma 1, the queue insertions and reinsertions (due to the minimum heap) would be finite. This leads to two possible termination cases.

\textbf{Case (i):} The goal $t$ is popped from the queue.  
Then $t \neq s$ must have been discovered from a neighbor $n$ and assigned
$\texttt{parent}(t) = n$. Recursively following parent pointers decreases $g$
monotonically and must reach $s$ in finitely many steps, yielding a valid path prioritizing safety (Lemma 2)).

\textbf{Case (ii):} The queue becomes empty without expanding $t$.  
Then all free states reachable from $s$ have been expanded or found not to
improve the cost. Hence, $t$ is unreachable via free states and no
collision-free path exists.

\medskip
In both cases the returned result is correct. Thus, UPP is complete.
\end{proof}

\subsubsection{Effect of adaptive $\beta$}
We analyze how the adaptive safety weight $\beta$ affects the heuristic
values along the path and consequently influences both suboptimality and
safety. For any node $n$ with goal $t$, the heuristic is
\begin{equation}
    h^{(\beta)}(n,t)
    = d(n,t) + \beta\, S(n),
\end{equation}
where $d(n,t)$ is the geometric distance term and $S(n)$ is the
proximity-based safety cost. Assume the successors 
$k_i, i=1,2....b$ are elements of the current OPEN set during expansion, and are from almost equal distances away from the goal.

Consider two safety weights $\beta_o$ (old) and $\beta_n$ (new).  
For any forward neighbor $k_{b+i}$, the change in heuristic becomes
\begin{equation}
    \delta h
    = h^{(\beta_n)}(k_{b+i}) - h^{(\beta_o)}(k_{b+i})
    = (\beta_n - \beta_o)\, S(k_{b+i}).
    \label{eq:deltah}
\end{equation}

% ------------------------------------------------------------
On decreasing $\beta$ , $\beta_n < \beta_o$, then from \eqref{eq:deltah} we have
\[
\delta h = (\beta_n - \beta_o)S(k_{b+i}) < 0,
\]
because $S(k_{l+i}) \ge 0$.  
Thus, the heuristic systematically decreases at every successor node:
\begin{equation*}
    h^{(\beta_n)}(k_{l+i}) < h^{(\beta_o)}(k_{l+i}).
\end{equation*}

Substituting this decrease into the standard A* inequality,
\[
d(m,t)+\beta_o S(m)
    \le h^*(m,t) + \Delta,
\]
on given \(\Delta\) represents the bounded heuristic cost due to the average safety cost term, and using $-\delta h > 0$. Adding $\delta h $ on both sides
\begin{equation*}
d(n,t)+\beta_o S(n) + \delta h
    \le h^*(n,t) + \Delta + \delta h.
\end{equation*}

\begin{equation*}
d(n,t)+\beta_n S(n) 
    \le h^*(n,t) + \Delta + \delta h.
\end{equation*}

Since $-\delta h > 0$, the suboptimality bound decreases by $\delta h$ which leads to improved cost for suboptimality

While on increasing  $\beta$ , safety weight increases preferring safer nodes for further planning according to lemma 2.

\subsubsection{Effect of adaptive $\alpha$}
On the given distance heuristic assuming constant safety,
\begin{equation}
    d(n,t)=\alpha(t)\,l_1(n,t)+(1-\alpha(t))\,l_{\infty}(n,t).
\end{equation}
After the adaptive update
\( \alpha\) changed to \(\alpha_c\)
the change in the heuristic is
\begin{align*}
    \Delta d(n,t)
    &=d^{(\alpha_c)}(n,t)-d^{(\alpha)}(n,t) \\
    &=\alpha_c l_1(n,t)+(1-\alpha_c)l_\infty(n,t)
      -\bigl[\alpha l_1(n,t)+(1-\alpha)l_\infty(n,t)\bigr] \\
    &= (\alpha_c-\alpha)\bigl(l_1(n,t)-l_\infty(n,t)\bigr) \\
    &= \delta \alpha\,\bigl(l_1(n,t)-l_\infty(n,t)\bigr).
\end{align*}

For \(\delta\alpha = \alpha_C-\alpha\)

Since on grid maps $l_1(n,t) > l_{\infty}(n,t)$, the sign of $\Delta d(n,t)$
is determined entirely by the sign of $\delta\alpha$:

\begin{itemize}
    \item \textbf{$\delta\alpha>0$ :}
    Since $l_1(n,t) > l_{\infty}(n,t)$, we obtain $\Delta d(n,t)>0$.
    The Manhattan component gains weight, increasing the cost of lateral or
    diagonal deviation. Hence the heuristic favours grid-aligned, 
    goal-directed motion, reinforcing a correct heading.

    \item \textbf{$\delta\alpha<0$ :}
    Then $\Delta d(n,t)<0$, reducing the influence of $l_1$ and making the
    heuristic more Chebyshev-like. This lowers directional bias and permits
    diagonal, exploratory adjustments, which is useful when the current
    motion is ineffective or moving away from the goal.
\end{itemize}

As the sign of $\Delta d(n,t)$ matches the sign of $\delta\alpha$,
adaptive $\alpha$ acts as a direction-aware shaping term: progress toward
the goal strengthens commitment to the current direction, while the regressor
reduces this commitment and promotes reorientation. This establishes a
self-correcting directional mechanism that improves convergence and avoids
persistent deviation from the goal.

\subsection{Algorithm}
The pseudo-algorithm for the UPP main search is presented in Algorithm \ref{alg:upp_main}. UPP computes a path over a grid from a given start to a given goal using adaptive heuristic components. The process begins with the geometry-aware initialization of parameters, described in Algorithm \ref{alg:upp_init}. Here, the safety weight $\beta$ is initialized based on obstacle density and spatial variability, while the radius $R$ is scaled according to the proximity of obstacles, providing an estimate of how much local risk needs to be evaluated. The safety field is computed as an inverse-distance accumulation over a Chebyshev neighborhood at this stage and serves as a safety-aware heuristic term.

During the graph search in Algorithm \ref{alg:upp_main}, the UPP continuously adapts its heuristic parameters using the update rules in Algorithm \ref{alg:upp_update}. The safety parameter $\beta$ is adjusted through stall detection when the search fails to progress meaningfully toward the goal, while the distance mixing parameter $\alpha$ is updated based on directional alignment with the goal. These adaptive updates enable UPP to dynamically balance optimality, safety, and smoothness throughout the planning process.

\begin{algorithm}[H]
\small
\caption{\textsc{InitUPP}: Geometry Aware Initialization}
\label{alg:upp_init}
\KwIn{Occupancy grid $G$; base $(\alpha_{\text{base}},\beta_{\text{base}},R_{\text{base}},\varepsilon)$}
\KwIn{Bounds $(\beta_{\min},\beta_{\max},R_{\min},R_{\max})$}
\KwOut{Initialized $\alpha,\beta,R$ and safety field $S(\cdot)$}

\BlankLine
$\alpha \gets \alpha_{\text{base}}$;\quad
$\beta \gets \beta_{\text{base}}$;\quad
$R \gets R_{\text{base}}$\;

Compute Euclidean distance transform $D$ over free cells of $G$\;
$\mathcal{F} \gets \{x \mid G(x)=0\}$;\quad
$\mathcal{O} \gets \{x \mid G(x)=1\}$\;
$\mu \gets \mathrm{mean}(D(x)\,|\,x\in\mathcal{F})$;\quad
$\sigma \gets \mathrm{std}(D(x)\,|\,x\in\mathcal{F})$;\quad
$\rho \gets \frac{|\mathcal{O}|}{|G|}$\;

$\beta \gets \mathrm{clip}\!\Big(\beta\,\rho\,\frac{\sigma}{\mu+\varepsilon},\,\beta_{\min},\beta_{\max}\Big)$\;
$R \gets \mathrm{clip}\!\big(\mathrm{round}(R(\mu+\sigma)),\,R_{\min},R_{\max}\big)$\;

\BlankLine

Build Chebyshev kernel $K(\Delta) = 1/(\|\Delta\|_\infty+\varepsilon)$ for $1 \le \|\Delta\|_\infty \le R$, else $0$\;
Convolve obstacle grid with $K$ to obtain $S_{\text{sum}}$\;
Set $S(x) \gets S_{\text{sum}}(x)$ for free cells, $0$ for obstacles\;

\Return $(\alpha,\beta,R,S(\cdot))$\;
\end{algorithm}

\begin{algorithm}[htpb]
\small
\caption{\textsc{UpdateParams}: Adaptive Update of $\beta$ and $\alpha$}
\label{alg:upp_update}

\KwIn{Current node $n_i$, goal $t$, parent map $\mathrm{parent}(\cdot)$}
\KwIn{Current parameters $\alpha_i,\beta_i$}
\KwIn{State variables: previous goal distance $d_{i-1}$, 
stall counter $s_{i-1}$, accumulated angular deviation $u$, 
and angular-window counter $q$}
\KwIn{$\beta$-update parameters:
$\tau_{\mathrm{goal}},K_{\beta},\beta_{\min},\beta_{\max},
\gamma_{\mathrm{dec}},\gamma_{\mathrm{rec}}$}
\KwIn{$\alpha$-update parameters:
$\tau_{\mathrm{ang}},K_{\alpha},\alpha_{\min},\alpha_{\max},
\eta_{\mathrm{dec}},\eta_{\mathrm{rec}},\theta_{\mathrm{tar}}$}

\KwOut{Updated $\alpha_{i+1},\beta_{i+1}$ and state variables 
$d_i,s_i,u,q$}

\BlankLine

\tcp{Compute signed progress toward the goal}
$d_i \gets \|n_i-t\|_2$\;
$\Delta d_i \gets d_i-d_{i-1}$\;

\BlankLine

\tcp{Compute progress indicator $p_i$}
\If{$\Delta d_i < -\tau_{\mathrm{goal}}$}{
    $p_i \gets +1$\tcp*[r]{progress}
}
\ElseIf{$\Delta d_i > \tau_{\mathrm{goal}}$}{
    $p_i \gets -1$\tcp*[r]{regression}
}
\Else{
    $p_i \gets 0$\tcp*[r]{stall}
}

\BlankLine

\tcp{Update stall counter and safety weight $\beta$}
\If{$p_i = +1$}{
    $s_i \gets 0$\;
    $\beta_{i+1}
      \gets
      \min(\beta_i\gamma_{\mathrm{rec}},\beta_{\max})$\;
}
\ElseIf{$p_i = -1$}{
    $s_i \gets 0$\;
    $\beta_{i+1}
      \gets
      \max(\beta_i\gamma_{\mathrm{dec}},\beta_{\min})$\;
}
\Else{
    $s_i \gets s_{i-1}+1$\;
    $\beta_{i+1} \gets \beta_i$\;

    \If{$s_i \ge K_{\beta}$}{
        $\beta_{i+1}
          \gets
          \max(\beta_i\gamma_{\mathrm{dec}},\beta_{\min})$\;
        $s_i \gets 0$\;
    }
}

\BlankLine

\tcp{Compute angular deviation $\theta_i$}
\If{$\mathrm{parent}(n_i)$ is defined}{
    $p \gets \mathrm{parent}(n_i)$\;
    
    $\mathbf{v}^{\mathrm{move}}_i \gets n_i-p$\;
    $\mathbf{v}^{\mathrm{goal}}_i \gets t-n_i$\;

    $\theta_i^{\mathrm{move}}
      \gets
      \operatorname{atan2}
      (v^{\mathrm{move}}_{i,y},v^{\mathrm{move}}_{i,x})$\;

    $\theta_i^{\mathrm{goal}}
      \gets
      \operatorname{atan2}
      (v^{\mathrm{goal}}_{i,y},v^{\mathrm{goal}}_{i,x})$\;

    $\theta_i
      \gets
      \left|
      \operatorname{wrap}_{[-\pi,\pi]}
      \left(
      \theta_i^{\mathrm{goal}}
      -
      \theta_i^{\mathrm{move}}
      \right)
      \right|$\;
}
\Else{
    $\theta_i \gets 0$\;
}

\BlankLine

\tcp{Accumulate turning deviation over a window of length $K_{\alpha}$}
$u \gets u+(\theta_i-\theta_{\mathrm{tar}})$\;
$q \gets q+1$\;

$\alpha_{i+1} \gets \alpha_i$\;

\If{$q \ge K_{\alpha}$}{
    \If{$u > \tau_{\mathrm{ang}}$}{
        $\alpha_{i+1}
          \gets
          \min(\alpha_i\eta_{\mathrm{rec}},\alpha_{\max})$\;
    }
    \ElseIf{$u < -\tau_{\mathrm{ang}}$}{
        $\alpha_{i+1}
          \gets
          \max(\alpha_i\eta_{\mathrm{dec}},\alpha_{\min})$\;
    }

    $u \gets 0$\;
    $q \gets 0$\;
}

\BlankLine

\Return $(\alpha_{i+1},\beta_{i+1},d_i,s_i,u,q)$\;

\end{algorithm}
\begin{algorithm}[H]
\small
\caption{Unified Path Planner (UPP): Main Search}
\label{alg:upp_main}
\KwIn{Occupancy grid $G$, start node $s$, goal node $t$}
\KwIn{Parameters:
$\{ \alpha_{\mathrm{base}}, \beta_{\mathrm{base}}, R_{\mathrm{base}}, \varepsilon,
\beta_{\min}, \beta_{\max}, \gamma_{\mathrm{dec}}, $\\$\gamma_{\mathrm{rec}}, K_{\beta}, \tau_{\mathrm{goal}},
\alpha_{\min}, \alpha_{\max}, \eta_{\mathrm{dec}}, \eta_{\mathrm{rec}}, \tau_{\mathrm{ang}},
\theta_{\mathrm{tar}}, K_{\alpha}, R_{\min}, R_{\max} \}$}
\KwOut{Path $\pi$ from $s$ to $t$ if found; failure otherwise}

\BlankLine

\If{$s$ or $t$ invalid or occupied in $G$}{\Return failure}
\If{$s = t$}{\Return $(s)$}

\BlankLine

$(\alpha,\beta,R,S) \gets \textsc{InitUPP}(G)$\;
Initialize $g(n) \gets +\infty$ for all $n$; $g(s) \gets 0$\;
$V \gets \emptyset$; $\mathrm{parent}(\cdot)$ undefined\;
Define $h(n,t) = \alpha\,\ell_1(n,t) + (1-\alpha)\,\ell_\infty(n,t) + \beta\,S(n)$\;
Initialize OPEN as priority queue; push $(f(s),g(s),s)$ with $f(s) = g(s) + h(s,t)$\;
$\text{prev\_dist} \gets \|s-t\|_2$;\quad
$\text{stalled} \gets 0$;\quad
$\text{turn\_sum} \gets 0$;\quad
$\text{turn\_iter} \gets 0$\;

\BlankLine

\While{OPEN not empty}{
    Pop $(f,g,n)$ with smallest $f$ from OPEN\;
    \If{$n \in V$}{\textbf{continue}}
    Add $n$ to $V$\;
    \If{$n = t$}{\textbf{goto} PathReconstruction}

    $(\alpha,\beta,\text{prev\_dist},\text{stalled},\text{turn\_sum},\text{turn\_iter}) \gets
    \textsc{UpdateParams}(n,t,\mathrm{parent},\alpha,\beta,\text{prev\_dist},\text{stalled},\text{turn\_sum},\text{turn\_iter})$\;

    \ForEach{neighbor $m \in \Gamma(n)$}{
        \If{$m$ invalid, occupied, or $m \in V$}{\textbf{continue}}
        Compute step cost $c_{nm}$ \;
        $g_{\text{new}} \gets g + c_{nm}$\;
        \If{$g_{\text{new}} < g(m)$}{
            $g(m) \gets g_{\text{new}}$;\quad $\mathrm{parent}(m) \gets n$\;
            $f(m) \gets g(m) + h(m,t)$\;
            Push $(f(m),g(m),m)$ into OPEN\;
        }
    }
}

\BlankLine
\textbf{PathReconstruction:}\;
\If{$n \neq t$}{\Return failure}
Initialize empty path $\pi$\;
\While{$n$ defined}{
    Prepend $n$ to $\pi$;\quad $n \gets \mathrm{parent}(n)$\;
}
\Return $\pi$\;
\end{algorithm}

\section{OptiSafe Index}
\label{sec:optisafe}

This section introduces the \textit{OptiSafe} index, a normalized metric designed to evaluate planning algorithms by jointly considering \emph{optimality} and \emph{safety}. Unlike metrics that assess these dimensions separately, the OptiSafe index penalizes imbalance towards one metric and rewards equal uplift. 

Consider a path planning algorithm $P$ given a start position $s$ and an end goal $ t$. We denote the path length of $P$ by $L(P)$. Correspondingly, we denote the optimal path length by $L(optimal)$. 

The deviation from optimality is defined as
\[
O_D = \min\!\left( 1,\; \max\!\left( 0,\; \frac{L(P)-L(\text{optimal})}{L(\text{optimal})}\right)\right).
\]

The \textit{optimality index} is defined as
\[
O(P) = 1 - O_D,
\]
which ensures $O(P)\in[0,1]$.

Additionally, we define the minimum clearance of the planner $P$ from the obstacles as $D(P)$. 
Consider a path planning algorithm that doesn't prioritize optimality but is tasked only with keeping the planner safe. The minimum clearance of the safe planner is denoted as $D(safe)$.

Define the \emph{safety deviation} as
\[
C_D =
\begin{cases}
0, & \text{if } D(P) \ge D(\text{safe}),\\[6pt]
\dfrac{D(\text{safe})-D(P)}{D(\text{safe})}, & \text{otherwise}.
\end{cases}
\]

The \textit{safety index} is
\[
C(P) = 1 - C_D,
\]
which also lies in $[0,1]$.

We combine the two into a balanced measure:
\begin{equation}
\mathrm{OptiSafe}(P) =
\Big(1-\lvert O(P)-C(P)\rvert\Big)\;
\frac{\sqrt{O(P)^2 + C(P)^2}}{\sqrt{2}}.
\label{eq:optisafe}
\end{equation}

For simplicity, set \(O:=O(P)\), \(C:=C(P)\). Then

\[
B(O,C) = 1 - |O-C| \quad \text{, balance term} \]
\[R(O,C) = \frac{\sqrt{O^2+C^2}}{\sqrt{2}} \quad \text{, strength term}.
\]

\begin{figure}[t]
    \centering
    \includegraphics[width=0.75\linewidth]{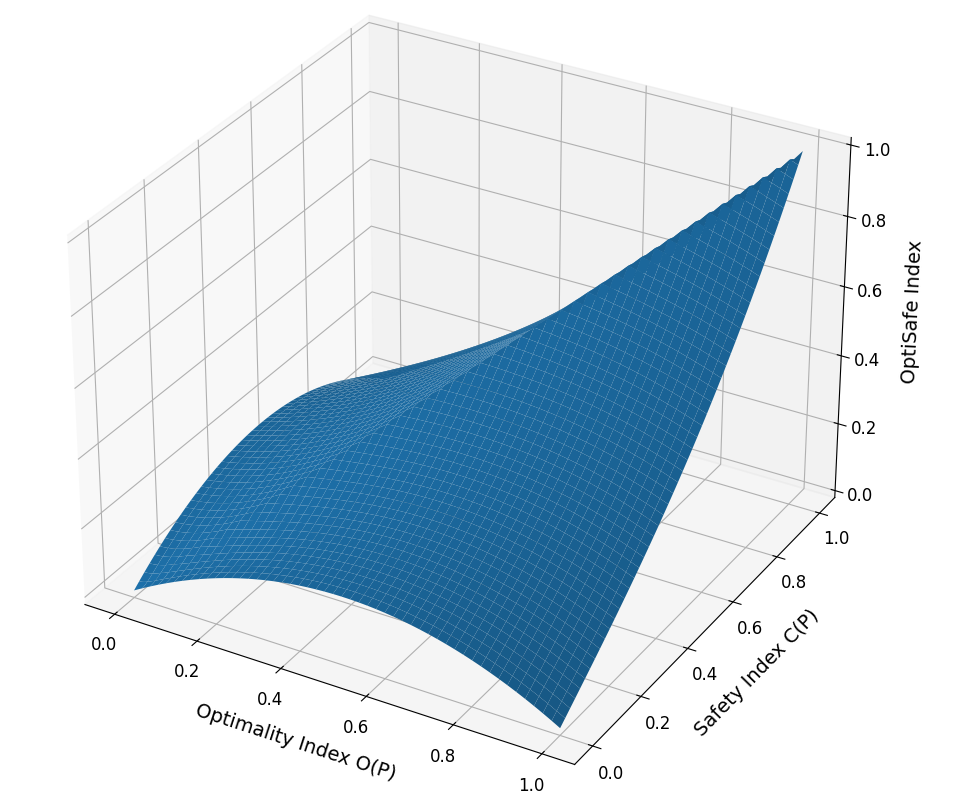}
    \caption{
        Optisafe Index Surface
    }
    \label{fig:Optisafe}
\end{figure}

An illustration of the Optisafe Index surface as a function of the optimality and safety indices is shown in Fig.~\ref{fig:Optisafe}. 

The OptiSafe Index is intended as a compact summary of the safety–optimality trade-off rather than a replacement for the individual performance measures. To preserve transparency, path length and minimum obstacle clearance are reported separately alongside the OptiSafe Index in all comparative evaluations. Therefore, the underlying optimality and safety behavior of each planner can be inspected directly even when planners obtain similar aggregate OptiSafe scores.

The formulation of the OptiSafe Index is motivated by two complementary 
requirements for evaluating safety-aware path planners. The balance term
$B(O,C) = 1 - |O-C|$ explicitly penalizes disparity between optimality 
and safety, attaining its maximum when the two normalized objectives are 
equal. However, balance alone is insufficient, since equally poor values 
of optimality and safety would also yield a maximum balance score. 
Therefore, the normalized magnitude term
$R(O,C) = \frac{\sqrt{O^2 + C^2}}{\sqrt{2}}$
measures the joint strength of the two objectives. The multiplicative 
combination $B(O,C)R(O,C)$ consequently requires a planner to exhibit 
both balanced and strong performance to obtain a high OptiSafe score: 
poor performance in either balance or joint strength directly reduces 
the overall score. The formulation is therefore a designed scalarization 
based on these desired properties rather than an arbitrary state-dependent penalty. Its boundedness, symmetry, and limiting behavior are established in the following theoretical analysis.

\subsection{Theoretical properties}
\label{subsec:optisafe_props}

\begin{lemma}\label{lem:O_in_01}
For all planners $P$, \(0 \le O(P) \le 1\) and \(0 \le C(P) \le 1\)
\end{lemma}
\begin{proof}
By construction, \(O_D=\min(1,\max(0,x))\) with \(x=\tfrac{L(P)-L(\text{optimal})}{L(\text{optimal})}\).
Thus \(0\le O_D\le 1\). Since \(O(P)=1-O_D\), we have \(0\le O(P)\le 1\).
Further for $C(P)$,

Case 1: If \(D(P)\ge D(\text{safe})\), then $C_D=0$ and $C(P)=1$.  
Case 2: If \(D(P)< D(\text{safe})\), then
$0 \le (D(\text{safe})-D(P))/D(\text{safe}) \le 1$. Hence $0\le C_D\le 1$, and $C(P)=1-C_D\in[0,1]$.
\end{proof}

\begin{lemma}\label{lem:balance_strength}
Considering $O(P)$ and $C(P)$ to be $O$ and $C$ respectively. As $O,C \in [0,1]$. Define
\[
B(O,C) = 1 - |O - C|,
\qquad
R(O,C) = \frac{\sqrt{O^2 + C^2}}{\sqrt{2}}.
\]
Then the following hold:
\begin{enumerate}
\item $0 \le B(O,C) \le 1$ and $0 \le R(O,C) \le 1$.
\item $B(O,C) = 1$ if and only if $O = C$, and $B(O,C) = 0$ if and only if $|O - C| = 1$.
\item $R(O,C) = 1$ if and only if $O = C = 1$, and $R(O,C) = 0$ if and only if $O = C = 0$.
\end{enumerate}
\end{lemma}

\begin{proof}
Since $O,C \in [0,1]$, it follows that $|O-C| \in [0,1]$, and hence
\[
0 \le B(O,C) = 1 - |O-C| \le 1.
\]
The equality cases follow directly from properties of the absolute value.

Moreover, $0 \le O^2 + C^2 \le 2$, which implies
\[
0 \le R(O,C) = \frac{\sqrt{O^2 + C^2}}{\sqrt{2}} \le 1.
\]
The upper bound is achieved if and only if $O^2 + C^2 = 2$, i.e., $O=C=1$, while the lower bound holds if and only if $O=C=0$.
\end{proof}

% \begin{theorem}\label{thm:range}
% For all $O,C\in[0,1]$,
% \[
% 0 \le \mathrm{OptiSafe}(O,C) \le 1.
% \]
% \end{theorem}
% \begin{proof}
% From lemma \ref{lem:balance_strength} , $B\in[0,1]$ and $R\in[0,1]$. Their product lies in $[0,1]$.  
% upper bound $1$ is achieved at $O=C=1$, lower bound $0$ occurs if $O=C=0$ or $|O-C|=1$.
% \end{proof}

\begin{theorem}
For all $O,C \in [0,1]$, 
\begin{enumerate}
\item \(
0 \le \mathrm{OptiSafe}(O,C) \le 1.
\)
\item \(\mathrm{OptiSafe}(O,C)=\mathrm{OptiSafe}(C,O)\).
   
\end{enumerate}
\end{theorem}
\begin{proof}

(i) From lemma \ref{lem:balance_strength} , $B\in[0,1]$ and $R\in[0,1]$. Their product lies in $[0,1]$.  
upper bound $1$ is achieved at $O=C=1$, lower bound $0$ occurs if $O=C=0$ or $|O-C|=1$.

(ii) Both components are symmetric: $|O-C|=|C-O|$, and $O^2+C^2=C^2+O^2$.
\end{proof}

\textit{Remark:} OptiSafe enforces \emph{conditional monotonicity}, rewarding equal uplift and penalizing imbalance. This design ensures that planners are evaluated not only for strength but also for harmony between optimality and safety.

\subsection{Algorithm}
\begin{algorithm}[H]
\label{alg:optisafe}
\caption{OptiSafe Index Evaluation}
\KwIn{Start state $s$, goal state $t$, grid map $G$, cell size $c$, planner path $P$}
\KwOut{OptiSafe Index $\mathcal{O}$}
\BlankLine

$P_{optimal} \leftarrow$ OptimalPlanner.plan($s,g,M$) \\
$L(Optimal) \leftarrow$ PathCost($P_{optimal},c$) \\
$L(P) \leftarrow$ PathCost($P,c$) \\

$P_{safe} \leftarrow$ HighSafetyPlanner.plan($s,g,M$) \\
$D(safe) \leftarrow$ MinimumClearance($P_{safe},c$) \\
$D(P) \leftarrow$ MinimumClearance($P,c$) \\
\eIf{$D(safe) > 0$}{
    \eIf{$D(P) > D(safe)$}{$cd \leftarrow 0$}{$cd \leftarrow \dfrac{D(safe) - D(P)}{D(safe)}$}
}{
    $cd \leftarrow 1$
}

\If{$L(Optimal) > 0$}{
    $od \leftarrow \min\!\Big(1, \max\!\big(\tfrac{L(P) - L(Optimal)}{L(Optimal)}, 0\big)\Big)$
}
\Else{$od \leftarrow 1$}

$o \leftarrow 1 - od$ \\
$c \leftarrow 1 - cd$ \\
$\mathcal{OSI} \leftarrow \dfrac{(1-|o-c|)\cdot\sqrt{o^2+c^2}}{\sqrt{2}}$ \\

\Return $\mathcal{OSI}$
\end{algorithm}

The pseudo-algorithm for computing the OptiSafe Index is presented in Algorithm \ref{alg:optisafe}. For the planner under evaluation, we first obtain its path using its standard planning routine. To normalize safety and optimality, we additionally compute two reference paths: (i) an optimal path, which provides the minimal achievable path length, and (ii) a safe path, which provides the maximal achievable clearance. These reference paths define the lower bound on path cost and the upper bound on safety, respectively. Using these values, the optimality and safety indices are computed as described earlier. Finally, substituting these indices into Eq. \ref{eq:optisafe}, the OptiSafe Index is evaluated.

\section{Simulation results}

\subsection{UPP ablation studies}
We perform ablation studies for UPP on a $1000 \times 1000$ grid with a cell size of $0.05\,\mathrm{m}$. All path lengths (in m) and clearances (in cm) are reported in the world frame with grid size taken into account. Table~\ref{tab:adaptivity} evaluates the effect of making the parameters $\alpha$ and $\beta$ adaptive in both a sparse and a cluttered environment. These simulations have been conducted on a 13th Gen Intel(R) Core(TM) i7-13650HX 2.60 GHz processor and 16 GB RAM.

In the sparse environment, using fixed $(\alpha,\beta)$ requires approximately $2204.69\,\mathrm{ms}$ of planning time. Making only $\alpha$ adaptive reduces the planning time to $332.50\,\mathrm{ms}$, reducing it drastically while balancing diagonal and grid-based movements, while keeping the path length essentially unchanged. When only $\beta$ is adaptive, the planning time remains comparable to the fixed case, but the minimum clearance increases from $0.14\,\mathrm{m}$ to $0.25\,\mathrm{m}$, reflecting more conservative, safety-oriented behavior in the low-density setting. When both $\alpha$ and $\beta$ are adaptive, planning time is further reduced to $267.06\,\mathrm{ms}$  and clearance increases to $0.30\,\mathrm{m}$. At the same time, the total turning angle drops from $228.82^\circ$ to $138.85^\circ$, indicating smoother, less zig–zag paths.

A similar pattern is observed in the cluttered environment. With both parameters fixed, UPP requires $8383.19\,\mathrm{ms}$ and produces high-angle paths with a total turning angle of $2794.59^\circ$. Making only $\alpha$ adaptive reduces planning time to $1381.68\,\mathrm{ms}$ and decreases the total turning angle to $769.76^\circ$, while keeping path length and clearance essentially unchanged. In contrast, making only $\beta$ adaptive slightly increases planning time and turning, and does not improve clearance in the cluttered case, as the planner tends to spend more time in high-safety regions and can get stuck in local safe pockets. When both parameters are adaptive, the behavior closely matches the adaptive-$\alpha$ case: planning time is $1431.35\,\mathrm{ms}$ and the turning angle is $769.76^\circ$, suggesting that adaptivity in $\alpha$ is primarily responsible for the efficiency and smoothness gains in cluttered environments.

\begin{table}[htbp]
\centering
\caption{Effect of adaptive $\alpha$ and $\beta$ in sparse and cluttered environments.}
\label{tab:adaptivity}
\begin{tabular}{lcccc}
\toprule
\multicolumn{5}{c}{\textbf{Sparse Environment}}\\
\midrule
Method & Time (ms) & Length (m) & Clearance (cm) & Turn (deg) \\
\midrule
Both fixed & 2204.69 & 75.27 & 14.21 & 228.82 \\
Adaptive $\alpha$ & 332.50 & 75.30 & 18.54 & 228.82 \\
Adaptive $\beta$ & 2228.83 & 75.41 & 25.14 & 228.84 \\
Both Adaptive & 267.06 & 75.44 & 30.12 & 138.85 \\
\midrule
\multicolumn{5}{c}{\textbf{Cluttered Environment}}\\
\midrule
Method & Time (ms) & Length (m) & Clearance (cm) & Turn (deg) \\
\midrule
Both fixed & 8383.19 & 73.32 & 25.47 & 2794.59 \\
Adaptive $\alpha$ & 1381.68 & 73.50 & 30.32 & 769.76 \\
Adaptive $\beta$ & 8502.23 & 73.32 & 25.75 & 2884.58 \\
Both Adaptive & 1431.35 & 73.51 & 30.59 & 769.76 \\
\bottomrule
\end{tabular}
\end{table}
Table~\ref{tab:robustness} studies the robustness of UPP with fully adaptive parameters to the initialization $(\alpha_0,\beta_0)$. We vary $(\alpha_0,\beta_0)$ over more than one order of magnitude in $\beta_0$, from $(0.25,2.5)$ to $(0.75,40)$. Across all tested initializations, path length, clearance, and turning angle are identical (up to numerical rounding) in both sparse and cluttered environments. Planning time varies by at most about $10\%$. These results indicate that UPP with adaptive parameters is insensitive to the choice of initial $(\alpha_0,\beta_0)$ and does not require careful tuning.

\begin{table}[htbp]
\centering
\caption{Robustness to initialization of $(\alpha_0,\beta_0)$ with full adaptivity.}
\label{tab:robustness}
\begin{tabular}{lcccc}
\toprule
\multicolumn{5}{c}{\textbf{Sparse Environment}}\\
\midrule
{Init $(\alpha_0,\beta_0)$} & Time (ms) & Length (m) & Clearance (cm) & Turn (deg) \\
\midrule
(0.25, 2.5) & 304.67 & 75.44 & 30.13 & 138.85 \\
(0.50, 10)  & 271.38 & 75.44 & 30.13 & 138.85 \\
(0.75, 40)  & 287.75 & 75.44 & 30.13 & 138.85 \\
\midrule
\multicolumn{5}{c}{\textbf{Cluttered Environment}}\\
\midrule
{Init $(\alpha_0,\beta_0)$} & Time (ms) & Length (m) & Clearance (cm) & Turn (deg) \\
\midrule
(0.25, 2.5) & 1440.98 & 73.51 & 30.13 & 769.76 \\
(0.50, 10)  & 1405.65 & 73.51 & 30.13 & 769.76 \\
(0.75, 40)  & 1466.70 & 73.51 & 30.13 & 769.76 \\
\bottomrule
\end{tabular}
\end{table}

\subsection{Planners Comparison}

The planners have been compared on various metrics such as planning time (Time (ms)), minimum Clearance along the path (Clearance (cm)), turning angle (Turn(deg)), path length (Length(m)) , success rate (SR, in percent) and OptiSafe Index(OSI). The comparison of UPP has been done with A* (\cite{Hart1968}),Voronoi Planner (\cite{Ayawli2019}), RRT (\cite{Lav06}), SDF-A* (\cite{Wang2023}), Optimized-A* (\cite{Chu2024}), CBF-RRT (\cite{Manjunath2021}), FS Planner\cite{Cobano2025}).They have been compared on a 1000 X 1000 grid, with a resolution of 0.05 m for 100 random starts and goals for all environments. Metrics results have been scaled according to the resolution. Planners have been compared in a sparse environment (Table \ref{tab:main_sim_results_env1}) and in a cluttered environment (Table \ref{tab:main_sim_results_env2}). Further simulation details and rigorous evaluations across different environments have been added to the supplementary materials. The code for planner comparison is available at \url{https://github.com/jatinarora30/safeplan}. The Nav2 version linked to compare on the actual robot package is available at
\url{https://github.com/jatinarora30/nav2_safeplan}.

\begin{table}[htbp]
\centering
\small
\caption{Planners comparison in Sparse Env (Map 1)}
\label{tab:main_sim_results_env1}
\begin{tabular}{lcccccc}
\toprule
Planner & Time (ms) & Clearance (cm) & Turn (deg) & Length (m) & SR (\%) & OSI \\
\midrule
Voronoi    & 23.63   & 99.06 & 74.68   & 33.39 & 84  & 0.583 \\
A*         & 124.01  & 25.47 & 352.88  & 33.45 & 100 & 0.271 \\
RRT        & 1129.63 & 76.06 & 3624.14 & 41.47 & 89  & 0.366 \\
CBF-RRT    & 897.02  & 71.77 & 3552.70 & 40.95 & 91  & 0.386 \\
FS Planner & 2878.66 & 94.48 & 343.39  & 33.25 & 100 & 0.587 \\
SDF-A*     & 5283.81 & 277.07 & 2610.31 & 38.45 & 100 & 0.790 \\
Opt-A*     & 391.41  & 51.39 & 73.49   & 35.92 & 100 & 0.386 \\
UPP (Ours) & 279.83  & 42.04 & 450.08  & 33.56 & 100 & 0.575 \\
\bottomrule
\end{tabular}
\end{table}

\begin{figure}[t]
    \centering
    \includegraphics[width=0.75\linewidth]{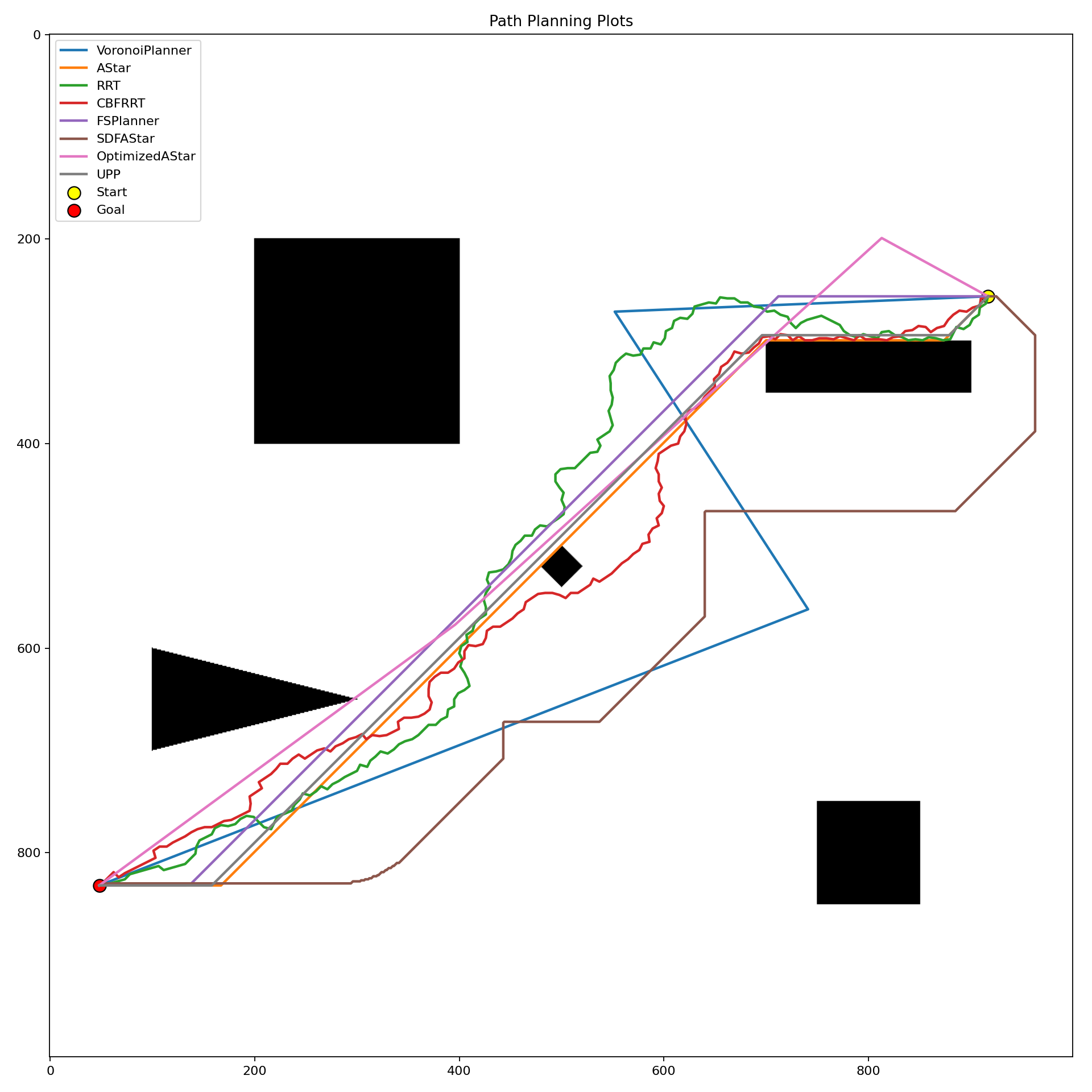}
    \caption{
        Planner comparison for random start and goal for Map 1
    }
    \label{fig:map1}
\end{figure}

On analyzing the results for the sparse environment (Table~\ref{tab:main_sim_results_env1}) and 
visualizing the generated paths (Fig.\ref{fig:map1}), several trends become evident. A* achieves a short 
path length (33.45\,m), but also exhibits the lowest minimum clearance among all planners, making it 
the least safe option. Conversely, SDF-A* produces the highest safety margin, although at the cost of 
approximately 15\% longer path length compared to the optimal. Thus, each method predominantly 
optimizes a single objective, either safety or optimality, without balancing the trade-off.

Voronoi planning provides both high clearance and near-optimal length, but its inability to reliably 
construct Voronoi diagrams in certain grid structures limits its practical applicability. RRT and 
CBF-RRT demonstrates high success rates; however, its sampling-based nature leads to extremely high 
turning angles and longer paths, indicating reduced smoothness and increased controller effort. 
Turning angle is an important metric for robot navigation because it directly reflects path 
smoothness and the ease of execution by a physical robot. FS Planner (FS) and Optimized A* 
(Opt-A*) performs similarly to UPP in terms of optimality with consistent safety margins. 
UPP achieves a path length of 33.56\,m, only a 0.5\% deviation from A* while maintaining higher 
clearance. Even though the clearance is significantly less than SDF-A*, FS, it is more computationally expensive than these algorithms. This demonstrates that 
UPP effectively balances optimality and safety at a considerably lower computational cost than other safety-aware planners.

To justify the effectiveness of the proposed OptiSafe Index (OSI), we analyze how it captures the 
joint balance between safety and optimality. As shown in Table~\ref{tab:main_sim_results_env1}, A* 
exhibits the lowest path length but a low OSI value of 0.271, reflecting its strong bias toward 
optimality with minimal emphasis on safety. Voronoi achieves a higher OSI of 0.583 because it 
maintains both good clearance and near-optimal length. RRT and CBF-RRT yield relatively low OSI 
values despite high minimum clearance, as their long and non-smooth paths significantly reduce their 
overall safety optimality balance.

SDF-A* obtains the highest OSI (0.790), owing to its consistently high-clearance paths with only 
moderate deviation (approximately 15\%) from the optimal path length. UPP achieves an OSI of 0.575, 
substantially higher than A* and comparable to FS Planner and SDF, maintaining near-optimal path length, at considerably low computation cost. This indicates that UPP offers one of the best practical 
trade-offs among the evaluated planners for practical uses, achieving a balanced combination of safety and optimality.

\begin{table}[htbp]
\centering
\caption{Planners comparison in Cluttered Env (Map 2)}
\label{tab:main_sim_results_env2}
\small
\begin{tabular}{lcccccc}
\toprule
Planner & Time (ms) & Clearance (cm) & Turn (deg) & Length (m) & SR (\%) & OSI \\
\midrule
Voronoi    & 49.33   & 0.20   & 2.57    & 0.51  & 2   & 0.02 \\
A*         & 643.09  & 6.88   & 264.80  & 39.29 & 100 & 0.22 \\
RRT        & 2365.59 & 7.27   & 4457.76 & 50.28 & 77  & 0.27 \\
CBF-RRT    & 2954.64 & 9.93   & 4138.33 & 48.09 & 74  & 0.29 \\
FS Planner & 4548.80 & 18.53  & 321.89  & 39.19 & 100 & 0.36 \\
SDF-A*     & 6410.51 & 118.42 & 1112.94 & 43.37 & 100 & 0.85 \\
Opt-A*     & 1418.49 & 35.10  & 161.90  & 45.27 & 100 & 0.38 \\
UPP (Ours) & 1158.61 & 25.44  & 433.57  & 39.62 & 100 & 0.94 \\
\bottomrule
\end{tabular}
\end{table}

\begin{figure}[t]
    \centering
    \includegraphics[width=0.75\linewidth]{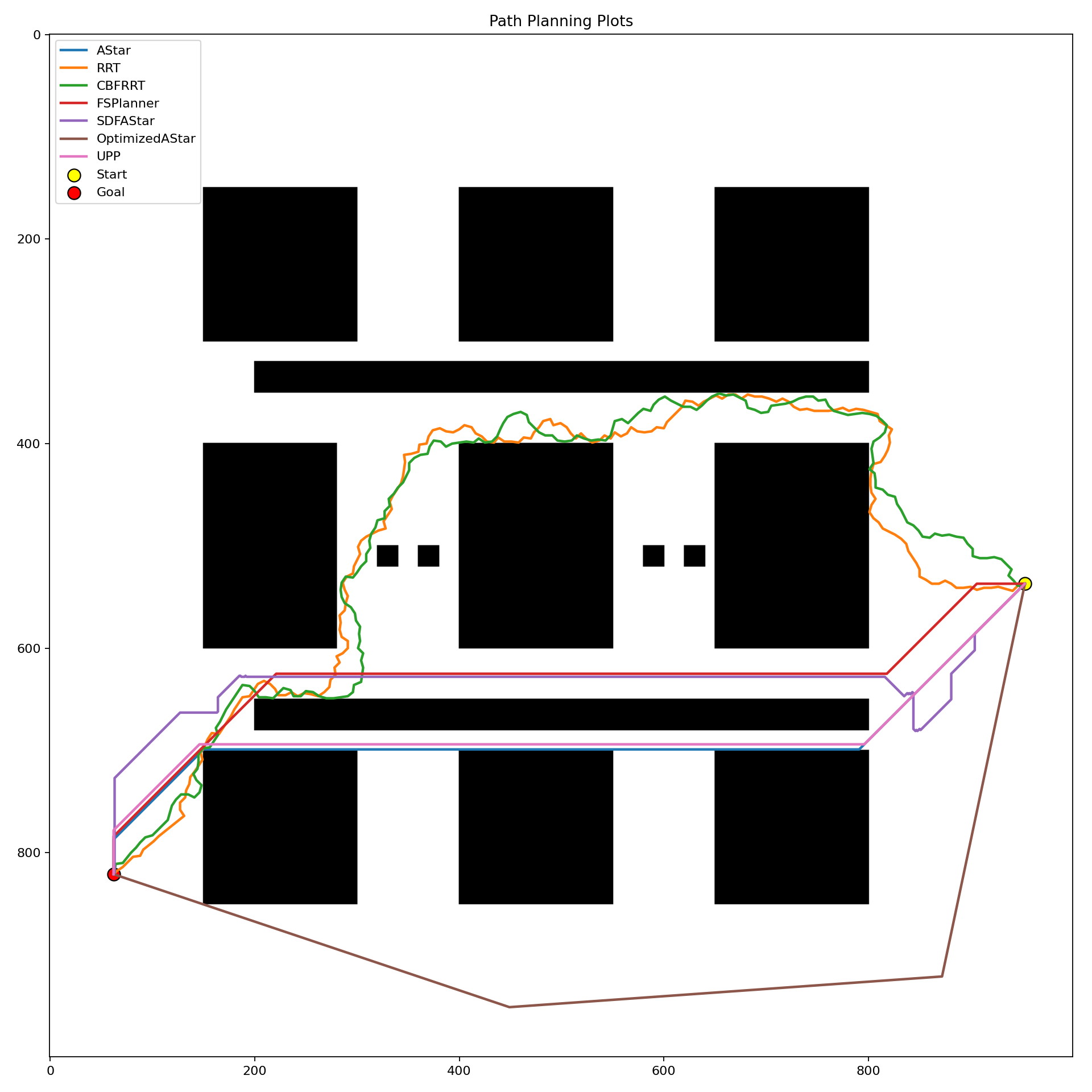}
    \caption{
        Planner comparison for random start and goal for Map 2
    }
    \label{fig:map2}
\end{figure}

On further analyzing the results for the cluttered environment (Table~\ref{tab:main_sim_results_env2}) 
and visualizing the generated paths (Fig.~\ref{fig:map2}), several important differences emerge 
compared to the sparse case. Voronoi performs poorly in this map, achieving a success rate of only 
2\% due to its inability to generate valid medial-axis structures in heavily cluttered regions. 
A* maintains low path length, but produces extremely low clearance (6.88 \,cm), indicating 
that it frequently drives through narrow passages with minimal safety margin.

Sampling-based planners (RRT and CBF-RRT) again show very high turning angles and long, non-smooth 
paths, which are undesirable for real robot execution. Their success rates also degrade 
substantially in this cluttered environment. SDF-A* achieves the highest clearance, but produces paths that are 
approximately 10\% longer than the optimal. In contrast, UPP maintains a high success rate and 
achieves a path length of 39.62\,m( approx 1\% optimal deviation) with a minimum clearance of 25.44\,cm, striking a strong balance 
between safety and path optimality. UPP's has the highest OSI score of 0.94 . This highlights UPP's robustness in cluttered 
scenarios, where it successfully avoids narrow unsafe regions while keeping the path length near 
optimal.

To further validate the OptiSafe Index (OSI) in the cluttered environment, we examine how it reflects 
the balance between safety and optimality across planners (Table~\ref{tab:main_sim_results_env2}). 
Voronoi achieves a very low OSI of 0.02, leading to unreliable behavior in densely cluttered maps. A* again shows a moderate OSI (0.22), capturing the fact that its paths exhibit extremely low clearance and therefore low safety. Sampling-based planners (RRT and CBF-RRT) also obtain low OSI values, since their high 
turning angles and long path lengths overshadow their minimal clearances, leading to poor safety-optimality trade-offs.

SDF-A* attains the second-highest OSI (0.85) by producing high-clearance paths with moderate deviation from 
optimal length, confirming its strong emphasis on safety in cluttered regions. UPP achieves the highest OSI of 
0.94,  maintaining significantly better near-optimal path length. This demonstrates that UPP consistently balances safety and optimality in a manner aligned with OSI, even in highly cluttered scenarios where traditional planners struggle to maintain both objectives simultaneously.

As shown in the tables, SDFA* and UPP attain comparable OptiSafe Index values in a sparse environment, with UPP requiring considerably less time. While UPP achieves the highest OptiSafe index in a cluttered environment, balancing both safety and optimality in a considerable manner.

The higher OptiSafe score of UPP compared with SDF-A*, despite the substantially larger raw clearance of SDF-A*, follows from the normalization used in the safety component. In our evaluation, the Voronoi planner is used as the high-safety reference planner for computing $D(\mathrm{safe})$. Consequently, once the clearance of an evaluated planner reaches or exceeds the corresponding Voronoi reference clearance, its safety index saturates at $C(P)=1$; additional clearance does not further increase the safety score. Thus, OptiSafe does not reward arbitrarily large obstacle clearance. In the cluttered environment, SDF-A* achieves very large clearance but also produces paths approximately $10\%$ longer than the near-optimal reference, which reduces its optimality index $O$. UPP, in contrast, generally satisfies the normalized safety requirement while maintaining a path length close to optimal. Hence, its normalized safety and optimality components are more closely balanced, resulting in a larger OptiSafe score.

The comparison with A* also clarifies the operational regime in which UPP
provides a distinct advantage. UPP is not intended to outperform A* on
pure path-length optimality or computational cost. In sparse environments,
where sufficient obstacle clearance is naturally available, A* remains a
strong choice due to its shortest-path behavior and lower planning time.
The advantage of UPP becomes more significant in cluttered environments,
where candidate paths with similar lengths can exhibit substantially
different obstacle clearances. In such cases, the adaptive safety term
allows UPP to favor higher-clearance alternatives while maintaining a
near-optimal path length. Therefore, UPP is particularly suited to
safety-sensitive navigation in cluttered environments, whereas standard A*
remains preferable when shortest-path optimality and computational
efficiency are the dominant requirements.

\section{Hardware experiments}

\subsection{Experimental setup}
In this section, we present the results of experiments on the proposed method using a hardware test bench for a TurtleBot robot. The TurtleBot was equipped with a LiDAR sensor for Simultaneous Localization and Mapping (SLAM), and ROS2-humble served as the interface. Mapping was done using the cartographer package in the lab, resulting in an area with few obstacles. The benchmark code was integrated as a global planner in the Nav2 package on the Turtlebot, and DWA was used as a local planner for obstacle avoidance.

\subsection{Results and analysis}

\begin{table}[htbp]
\centering
\small
\caption{Average performance over 5 runs.}
\label{tab:avg_results}
\begin{tabular}{lccc}
\toprule
Planner & Length (m) & Clearance (cm) & Turn (deg) \\
\midrule
UPP (Ours) & 3.75 & 6.68 & 341.7 \\
FS Planner & 3.63 & 6.44 & 365.6 \\
SDF-A*     & 3.59 & 6.12 & 387.8 \\
\bottomrule
\end{tabular}
\end{table}
Table~\ref{tab:avg_results} summarizes the average path length, minimum obstacle clearance, and total turning angle over five runs for UPP and the compared planners. For experimentation purposes, top-performing safety planners have been evaluated. The hardware results reveal a performance gap compared to the simulation. 
UPP's path length (3.75m) exceeds both FSPlanner and SDF-A* by 3-4\%,  while simulation showed <1\% optimality deviation. We consider the sim-to-real gap a priority for possible future work. However, UPP consistently maintains greater minimum obstacle clearance and lower turning effort, resulting in safer and smoother trajectories. This behavior aligns with UPP's objective to prioritize safety-aware navigation while allowing a controlled deviation from the shortest path. Notably, the increase in path length remains marginal relative to the gains in safety and smoothness, indicating a favorable trade-off for navigation in cluttered environments where conservative motion is desirable.

\section{Conclusions, Future Work and Limitations}
In this paper, we have proposed a Unified Path Planner, a unified framework for optimality and safety in planning, compared to other algorithms that primarily focus on either optimality or safety. The flexibility in UPP allows auto-tuning at runtime based on the path's direction towards the goal and stalled safety. The bounded heuristic cost is a function of dimension, increasing with higher dimensions.The present study considers deterministic occupancy maps and does not
explicitly model sensing, localization, or motion uncertainty; incorporating
uncertainty-aware or application-specific clearance thresholds is therefore
an important direction for future work. Future work could include reducing computational cost in higher dimensions and improving UPP's ability to handle dynamic obstacles. We have also presented the OptiSafe Index, a normalized score that defines the balance and strength between optimality and safety. Extensive simulations and experiments have done for planners to evaluate across various metrics. It has been observed that UPP was able to balance optimality and safety in various environments.
\backmatter

\bmhead{Supplementary information}

Simulation plots and envs and parameters have been provided in the form of supplementary material with this paper.

\section*{Declarations}

%Some journals require declarations to be submitted in a standardised format. Please check the Instructions for Authors of the journal to which you are submitting to see if you need to complete this section. If yes, your manuscript must contain the following sections under the heading `Declarations':

\begin{itemize}
\item Funding
Not applicable

\item Conflict of interest/Competing interests

No conflict of interest/Competing interests between authors.

\item Ethics approval and consent to participate

Not applicable
\item Consent for publication

All authors have provided their consent for publication of this manuscript.

\item Data availability 

Not applicable
\item Code availability 

The link for code used in this study is available in manuscript.

\item Author contribution
Jatin Kumar Arora conceived the core idea,  designed and conducted the simulations and experiments, and drafted the manuscript. Soutrik Bandyopadhyay contributed to the theoretical formulation, analysis of the algorithm, and refinement of the methodology. Sunil Sulania assisted with the implementation and writing of the manuscript. Shubhendu Bhasin supervised the research, provided critical insights on the problem formulation and results, and contributed to manuscript revision. All authors reviewed and approved the final version of the manuscript.
\end{itemize}


\begin{thebibliography}{99}

\bibitem[Paden et al.(2016)]{Paden2016}
Paden, B., Cap, M., Yong, S. Z., Yershov, D. \& Frazzoli, E.,
``A survey of motion planning and control techniques for self-driving urban vehicles,''
\textit{IEEE Transactions on Intelligent Vehicles}, vol. 1, no. 1, pp. 33--55, 2016.

\bibitem[Li et al.(2015)]{ChaochengLi2015}
Li, C., Wang, J., Wang, X. \& Zhang, Y.,
``A model based path planning algorithm for self-driving cars in dynamic environment,''
\textit{Proceedings of the Chinese Automation Congress (CAC)}, pp. 1123--1128, 2015.

\bibitem[Lee et al.(2014)]{Lee2014}
Lee, U., Yoon, S., Shim, H., Vasseur, P. \& Demonceaux, C.,
``Local path planning in a complex environment for self-driving car,''
\textit{Proceedings of the IEEE Cyber Technology Conference}, pp. 445--450, 2014.

\bibitem[Dai et al.(2022)]{Dai2022}
Dai, Y., Xiang, C., Zhang, Y., Jiang, Y., Qu, W. \& Zhang, Q.,
``A review of spatial robotic arm trajectory planning,''
\textit{Aerospace}, vol. 9, no. 7, p. 361, 2022.

\bibitem[Korayem et al.(2011)]{Korayem2011}
Korayem, M. H., Nohooji, H. R. \& Nikoobin, A.,
``Path planning of mobile elastic robotic arms by indirect approach of optimal control,''
\textit{International Journal of Advanced Robotic Systems}, vol. 8, no. 1, 2011.

\bibitem[Kim and Shin(1985)]{Kim1985}
Kim, B. K. \& Shin, K. G.,
``Minimum-time path planning for robot arms and their dynamics,''
\textit{IEEE Transactions on Systems, Man, and Cybernetics}, vol. SMC-15, no. 2, pp. 213--223, 1985.

\bibitem[Klanke et al.(2006)]{Klanke2006}
Klanke, S., Lebedev, D., Haschke, R., Steil, J. \& Ritter, H.,
``Dynamic path planning for a 7-DOF robot arm,''
\textit{Proceedings of the IEEE/RSJ International Conference on Intelligent Robots and Systems (IROS)}, pp. 3879--3884, 2006.

\bibitem[Bortoff(2000)]{Bortoff2000}
Bortoff, S. A.,
``Path planning for UAVs,''
\textit{Proceedings of the American Control Conference}, vol. 1, pp. 364--368, 2000.

\bibitem[Hvezda et al.(2018)]{Hvezda2018}
Hvezda, J., Rybecky, T., Kulich, M. \& Preucil, L.,
``Context-aware route planning for automated warehouses,''
\textit{Proceedings of the IEEE Intelligent Transportation Systems Conference (ITSC)}, pp. 2955--2960, 2018.

\bibitem[Vivaldini et al.(2010)]{Vivaldini2010}
Vivaldini, K. C. T., Galdames, J. P. M., Bueno, T. S., Araujo, R. C., Sobral, R. M., Becker, M. \& Caurin, G. A. P.,
``Robotic forklifts for intelligent warehouses: Routing, path planning, and auto-localization,''
\textit{Proceedings of the IEEE International Conference on Industrial Technology (ICIT)}, pp. 1463--1468, 2010.

\bibitem[Warren(1993)]{Warren1993}
Warren, C. W.,
``Fast path planning using modified A* method,''
\textit{Proceedings of the IEEE International Conference on Robotics and Automation (ICRA)}, pp. 662--667, 1993.

\bibitem[Nash and Koenig(2013)]{Nash2013}
Nash, A. \& Koenig, S.,
``Any-angle path planning,''
\textit{AI Magazine}, vol. 34, no. 4, pp. 85--107, 2013.

\bibitem[Stentz(1994)]{Stentz1994}
Stentz, A.,
``Optimal and efficient path planning for partially-known environments,''
\textit{Proceedings of the IEEE International Conference on Robotics and Automation (ICRA)}, pp. 3310--3317, 1994.

\bibitem[Noreen et al.(2016)]{Noreen2016}
Noreen, I., Khan, A. \& Habib, Z.,
``Optimal path planning using RRT* based approaches: A survey,''
\textit{International Journal of Advanced Computer Science and Applications}, vol. 7, no. 11, 2016.

\bibitem[Nasir et al.(2013)]{Nasir2013}
Nasir, J., Islam, F., Malik, U., Ayaz, Y., Hasan, O., Khan, M. \& Muhammad, M. S.,
``RRT*-SMART: A rapid convergence implementation of RRT*,''
\textit{International Journal of Advanced Robotic Systems}, vol. 10, no. 7, 2013.

\bibitem[Wu et al.(2021)]{Wu2021}
Wu, Z., Meng, Z., Zhao, W. \& Wu, Z.,
``Fast-RRT: A RRT-based optimal path finding method,''
\textit{Applied Sciences}, vol. 11, no. 24, p. 11777, 2021.

\bibitem[LaValle(2006)]{Lav06}
LaValle, S. M.,
\textit{Planning Algorithms},
Cambridge University Press, 2006.

\bibitem[Jafarzadeh and Fleming(2018)]{Jafarzadeh2018}
Jafarzadeh, H. \& Fleming, C.,
``An exact geometry-based algorithm for path planning,''
\textit{International Journal of Applied Mathematics and Computer Science}, vol. 28, no. 3, pp. 493--504, 2018.

\bibitem[Ayawli et al.(2019)]{Ayawli2019}
Ayawli, B. B. K., Mei, X., Shen, M., Appiah, A. Y. \& Kyeremeh, F.,
``Mobile robot path planning in dynamic environment using Voronoi diagram and computational geometry technique,''
\textit{IEEE Access}, vol. 7, pp. 86026--86040, 2019.

\bibitem[Hwang and Ahuja(1992)]{Hwang1992}
Hwang, Y. K. \& Ahuja, N.,
``A potential field approach to path planning,''
\textit{IEEE Transactions on Robotics and Automation}, vol. 8, no. 1, pp. 23--32, 1992.

\bibitem[Rostami et al.(2019)]{Rostami2019}
Rostami, S. M. H., Sangaiah, A. K., Wang, J. \& Liu, X.,
``Obstacle avoidance of mobile robots using modified artificial potential field algorithm,''
\textit{EURASIP Journal on Wireless Communications and Networking}, 2019.

\bibitem[Azzabi and Nouri(2019)]{Azzabi2019}
Azzabi, A. \& Nouri, K.,
``An advanced potential field method proposed for mobile robot path planning,''
\textit{Transactions of the Institute of Measurement and Control}, vol. 41, no. 11, pp. 3132--3144, 2019.

\bibitem[Chu et al.(2024)]{Chu2024}
Chu, L., Wang, Y., Li, S., Guo, Z., Du, W., Li, J. \& Jiang, Z.,
``Intelligent vehicle path planning based on optimized A* algorithm,''
\textit{Sensors}, vol. 24, no. 10, p. 3149, 2024.

\bibitem[Manjunath and Nguyen(2021)]{Manjunath2021}
Manjunath, A. \& Nguyen, Q.,
``Safe and robust motion planning for dynamic robotics via control barrier functions,''
\textit{Proceedings of the IEEE Conference on Decision and Control (CDC)}, pp. 2122--2128, 2021.

\bibitem[Wang et al.(2023)]{Wang2023}
Wang, H., Lin, Y., Zhang, W., Ye, W., Zhang, M. \& Dong, X.,
``Safe autonomous exploration and adaptive path planning strategy using signed distance field,''
\textit{IEEE Access}, vol. 11, pp. 144663--144675, 2023.

\bibitem[Cobano et al.(2025)]{Cobano2025}
Cobano, J. A., Merino, L. \& Caballero, F.,
``Exploiting Euclidean distance field properties for fast and safe 3D planning with a modified Lazy Theta*,''
\textit{arXiv preprint arXiv:2505.24024}, 2025.

\bibitem[Hart et al.(1968)]{Hart1968}
Hart, P., Nilsson, N. \& Raphael, B.,
``A formal basis for the heuristic determination of minimum cost paths,''
\textit{IEEE Transactions on Systems Science and Cybernetics}, vol. 4, no. 2, pp. 100--107, 1968.

\end{thebibliography}
\end{document}